%% file: main.tex
\documentclass[letterpaper]{article} 
\usepackage[preprint]{aaai2027}  
\usepackage[hyphens]{url}  
\usepackage{graphicx} 
\usepackage{natbib}  
\usepackage{caption} 
\AtBeginEnvironment{thebibliography}{\raggedright}
\usepackage{algorithm}
\usepackage{algorithmic}
\usepackage{amsmath}
\usepackage{makecell}
\usepackage{array}
\usepackage{tabularx}
\usepackage{fontawesome5}

\usepackage{newfloat}
\usepackage{listings}
\usepackage[most]{tcolorbox} 
\newtcolorbox{takeawaybox}{ colback=blue!5, colframe=blue!20, boxrule=0.4pt, arc=2pt, left=6pt, right=6pt, top=4pt, bottom=4pt, before skip=6pt, after skip=6pt }

\DeclareCaptionStyle{ruled}{labelfont=normalfont,labelsep=colon,strut=off} 
\floatstyle{ruled}
\newfloat{listing}{tb}{lst}{}
\floatname{listing}{Listing}

\usepackage{booktabs}

\definecolor{stepfunblue}{RGB}{1,132,255}
\newcommand{\stepfunlogo}{%
    \begingroup
    \raisebox{0pt}{%
        \includegraphics[height=0.405in]{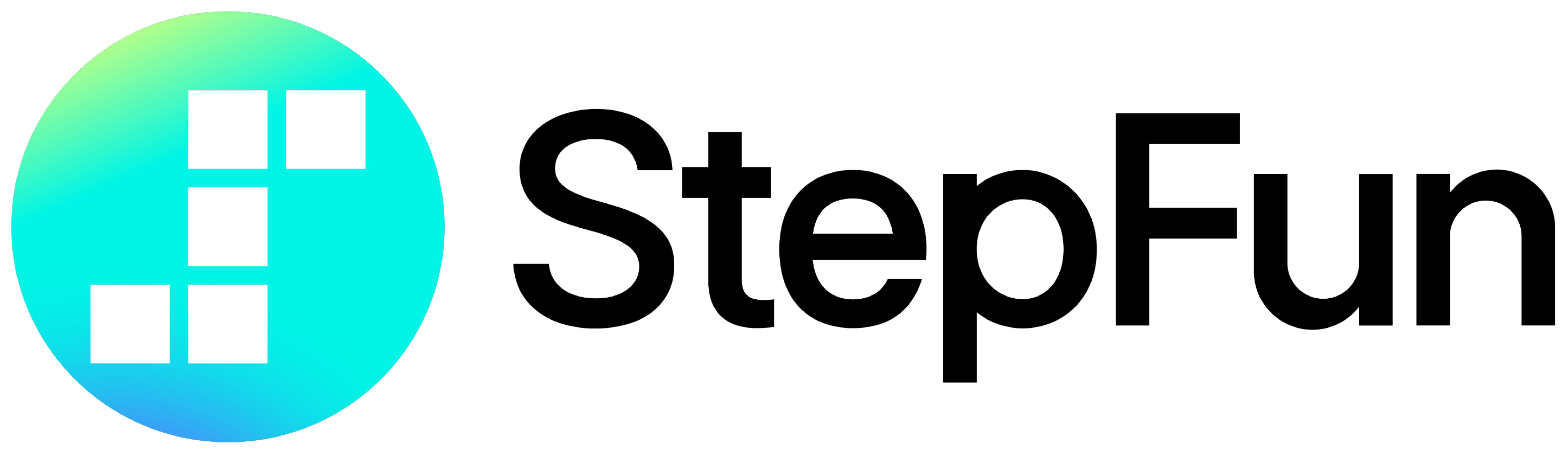}%
    }%
    \endgroup
}

\newcommand{\nativehref}[2]{%
    \begingroup
    \edef\linktarget{\detokenize{#1}}%
    \leavevmode
    \pdfstartlink
        attr{/Border [0 0 0]}
        user{/Subtype /Link /A << /S /URI /URI (\linktarget) >>}%
    #2%
    \pdfendlink
    \endgroup
}

\newlength{\stepfuntitletopskip}
\makeatletter
\xpatchcmd{\@maketitle}
    {\vskip 0.625in minus 0.125in}
    {\vskip\stepfuntitletopskip}
    {}{}
\xpatchcmd{\@maketitle}
    {\vskip 0.625in minus 0.125in}
    {\vskip\stepfuntitletopskip}
    {}{}
\makeatother

\title{%
    \makebox[\textwidth][l]{%
        \stepfunlogo
    }\par
    \vspace{0.04in}%
    \rule{\textwidth}{0.6pt}\par
    \vspace{0.12in}%
    $\Phi$-Bench: Can Large Language Models Engineer the Infrastructure
    \\That Powers Them?%
}
\author{
    \normalsize
    Leilei Ding\textsuperscript{* 1 \textdaggerdbl},
    Shumin Wang\textsuperscript{* 1},
    Yuting Huang\textsuperscript{* 1},
    Fanqi Wan\textsuperscript{2},
    Yinmin Zhang\textsuperscript{2},
    Qi Han\textsuperscript{2},
    Yiming Xu\textsuperscript{3},
    \\[0.5em]
    \normalsize
    Feiyuan Zhang\textsuperscript{4},
    Xiaomeng Chu\textsuperscript{5},
    Guoliang You\textsuperscript{6},
    Wuyang Zhang\textsuperscript{1},
    Daxin Jiang\textsuperscript{\textdagger \ \  2},
    Yanyong Zhang\textsuperscript{\textdagger \ \  1}
}

\affiliations{
    \small
    \textsuperscript{1}
    University of Science and Technology of China
    \\[0.5em]
    \small
    \textsuperscript{2}
    StepFun
    \\[0.5em]
    \small
    \textsuperscript{3}
    Peking University
    \\[0.5em]
    \small
    \textsuperscript{4}
    The Hong Kong University of Science and Technology
    \\[0.5em]
    \small
    \textsuperscript{5}
    Yale University
    \\[0.5em]
    \small
    \textsuperscript{6}
    University of Pennsylvania
    \\[0.7em]
    \normalsize
    \nativehref{https://faibench.org}{%
        \textcolor{stepfunblue}{\faGlobe\ \texttt{Leaderboard}}%
    }
    \qquad
    \nativehref{https://github.com/one2piece2hello/faibench_LLM_Infra_Bench}{%
        \faGithub\ \textcolor{stepfunblue}{\texttt{GitHub}}%
    }
    \qquad
    \nativehref{https://huggingface.co/datasets/faibench-Frontier-Infra-Bench/faibench_Frontier_Infra_Bench}{%
        \raisebox{-0.25em}{%
            \includegraphics[height=1.3em]{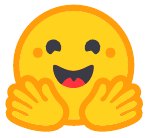}%
        }\
        \textcolor{stepfunblue}{\texttt{Hugging Face}}%
    }
}

\begin{document}

\maketitle

\begingroup
\renewcommand{\thefootnote}{\textdaggerdbl}
\footnotetext{Work done as an intern at StepFun.}
\renewcommand{\thefootnote}{*}
\footnotetext{Equal contribution.}

\renewcommand{\thefootnote}{\textdagger}
\footnotetext{
Correspondence to:
Daxin Jiang \texttt{<djiang@stepfun.com>} and Yanyong Zhang \texttt{<yanyongz@ustc.edu.cn>}.
}
\endgroup

\input{sections/0_abstract}
\input{sections/1_introduction}
\input{sections/2_related_work}
\input{sections/3_dataset}
\input{sections/4_evaluation}
\input{sections/5_discussion}
\input{sections/7_case}
\input{sections/6_conclusion}

\bibliography{aaai2027}
\input{sections/8_appendix.tex}


\end{document}

%% file: sections/0_abstract.tex
\begin{abstract}
Large language models (LLMs) now show strong reasoning and coding abilities, raising the prospect that they could assist in developing and optimizing the very infrastructure that powers them. Existing benchmarks, however, mainly focus on isolated kernels, predefined operators, or pre-specified optimization targets, and cannot evaluate open-ended, long-horizon LLM infrastructure engineering.
To address this gap, we present $\Phi$-Bench, a benchmark of 85 tasks grounded in real LLM training and inference codebases. 
The tasks cover nine categories of LLM infrastructure and come in three formats, ranging from kernel function completion to long-horizon implementation and repository-wide end-to-end optimization. They are built by a construction pipeline that turns real pull requests, issues, and repository code into executable tasks with validated tests, and every reward is gated by correctness and anti-hacking checks.
Across eight frontier models, the strongest model, Claude Opus 5, scores only 36.53\%, and every model performs worse on long-horizon, repository-level implementation than on kernel-level tasks.
Trajectory analysis further suggests that stronger models treat optimization as controlled experimentation, validating ideas with cheap local runs and ruling out confounders before drawing conclusions.
We will open-source $\Phi$-Bench to facilitate future research on LLM agents' ability to develop and optimize the LLM infrastructure that powers themselves.

\end{abstract}

%% file: sections/1_introduction.tex
\section{Introduction}

With recent advances in large language models (LLMs) in reasoning~\citep{guo2025deepseek} and code generation~\citep{glm5team2026glm5vibecodingagentic,openai2026gpt56}, leveraging LLMs to support the development of next-generation AI models has emerged as a promising research direction~\citep{ma2026trexautomatingllmfinetuning,ishibashi2025largelanguagemodelsinvent}. A particularly important challenge in this context is engineering and optimizing the software infrastructure underlying LLM training and inference, hereafter referred to as \emph{LLM infrastructure}. Prior work~\citep{kwon2023efficient,narayanan2021efficient} has demonstrated that infrastructure-level optimizations can substantially improve GPU utilization and reduce computational costs. These developments naturally raise an intriguing question: \textit{Can large language models engineer and optimize the infrastructure that powers them?}

Although several benchmarks evaluate LLMs on GPU kernel implementation~\citep{ouyang2025kernelbench,li2025tritonbench} and optimization~\citep{li2026cudahercules}, they typically focus on individual functions or small sets of isolated components, failing to capture the complexity of engineering and optimizing real-world LLM infrastructure. In practice, such work is long-horizon and open-ended: developers must navigate an existing infrastructure stack, identify bottlenecks, and iteratively implement, profile, debug, and refine solutions—an end-to-end process far beyond code completion or single-commit edits. Moreover, existing benchmarks cover only a narrow subset of LLM infrastructure engineering topics, limiting their ability to comprehensively assess whether LLMs can develop and optimize modern AI infrastructure. These gaps motivate a comprehensive, long-horizon, open-ended benchmark that faithfully evaluates whether LLMs can engineer the frontier LLM infrastructure.

\begin{figure*}[htb]
    \centering
    \includegraphics[width=0.99\linewidth]{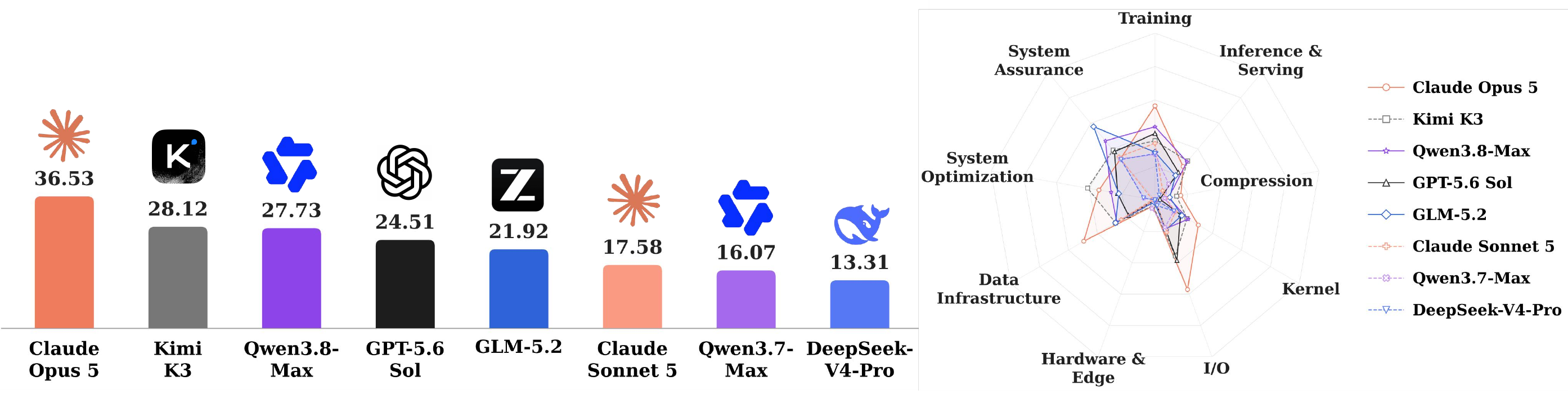}
    \caption{Performance of frontier models on $\Phi$-Bench. \textbf{Left}: Overall scores achieved by different models. \textbf{Right}: Model performance across different infrastructure topics.}
    \label{fig:main}
\end{figure*}

To bridge this gap, we introduce $\Phi$-Bench (the Frontier AI Infrastructure Benchmark), which draws on recent systems research and established open-source LLM infrastructure repositories to identify realistic, long-horizon and open-ended workloads for the evaluation of frontier LLM agents. Unlike benchmarks centered only on completing isolated functions, $\Phi$-Bench allows agents to navigate existing codebases and iteratively implement, profile, debug, and optimize their solutions. Constructing such a benchmark poses two challenges. First, LLM infrastructure spans heterogeneous areas of training and inference, so a task set drawn from a narrow class of components cannot represent the field. Second, scaling task construction is difficult: repository histories rarely provide ready-made benchmark tasks, so each task requires identifying a substantive infrastructure problem, locating its implementation, and reconstructing an executable, verifiable evaluation.

To address coverage, we analyze 2,260 systems papers and 1,852 public LLM infrastructure repositories to build a bottom-up taxonomy of 9 domains, 62 topics, and 410 tags. To scale construction, we synthesize tasks from historical issues and pull requests, code masking or infilling, and expert-curated research problems, with an iterative agent loop proposing tests for uncovered execution paths. From 5,215 candidates generated at substantial token and GPU-hour cost, automated validity checks and expert screening, auditing, and rewriting yield 85 executable tasks.

The resulting benchmark comprises 85 tasks in three formats of increasing scope and open-endedness: Kernel Function Completion (KFC), Long-Horizon Implementation (LHI), and End-to-End Optimization (E2EO). These formats graduate from local kernel implementation to repository-scale infrastructure development and end-to-end system optimization, enabling $\Phi$-Bench to distinguish agents' abilities in computational primitives, long-horizon codebase engineering, and hypothesis-driven optimization.

Using $\Phi$-Bench, we systematically evaluate frontier LLMs with low run-to-run variability. The best model, Claude Opus 5, scores \(36.53\%\), leaving substantial headroom. Additional experiments examine refinement iterations and reasoning budget, while trajectory analyses reveal distinct model strengths and weaknesses and highlight key challenges before LLMs can reliably engineer and optimize the infrastructure that powers them.

In summary, we make the following contributions:
\begin{itemize}
    \item We introduce \textbf{$\Phi$-Bench}, a comprehensive benchmark for evaluating LLM agents on realistic, long-horizon, and open-ended LLM infrastructure engineering.
    \item We develop a scalable task-construction framework that combines a bottom-up taxonomy with automated task synthesis, coverage-driven test generation, and expert validation.
    \item We systematically evaluate frontier LLMs on $\Phi$-Bench, revealing substantial headroom and recurring weaknesses in repository-level implementation and end-to-end optimization.
\end{itemize}

%% file: sections/2_related_work.tex
\section{Related Works}

\paragraph{LLM Infrastructure Optimization}

LLM infrastructure spans GPU computation and communication, data and I/O systems, and hardware-aware optimization across training and inference. CUTLASS and Triton support kernel development~\citep{Thakkar_CUTLASS_2023,tillet2019triton}, while FlashAttention and FlashInfer optimize attention computation~\citep{dao2022flashattention,ye2025flashinfer}. At the system level, Megatron-LM and DeepSpeed support scalable training~\citep{shoeybi2019megatron,rasley2020deepspeed}, whereas Orca, vLLM, and SGLang address serving concerns such as batching, scheduling, and KV-cache management~\citep{yu2022orca,kwon2023efficient,zheng2024sglang}.



\paragraph{Benchmarks for LLM Infrastructure Engineering}
Recent benchmarks have evaluated the ability of LLMs to implement and optimize components of LLM infrastructure~\citep{wang2026kernelbenchx,kernelbenchhard}. KernelBench~\citep{ouyang2025kernelbench}, TritonBench~\citep{li2025tritonbench}, and FlashInfer-Bench~\citep{xing2026flashinfer} focus on GPU operators with predefined interfaces, specifications, and objectives. They therefore avoid repository navigation, bottleneck identification, and coordinated changes across the software stack. ISO-Bench~\citep{nangia2026iso} and CUDAHercules~\citep{li2026cudahercules} extend evaluation to repository-level GPU optimization but typically identify the target component and bottleneck, limiting their ability to assess open-ended infrastructure engineering.




%% file: sections/3_dataset.tex
\begin{figure*}[!t]
    \centering
    \includegraphics[width=0.90\linewidth]{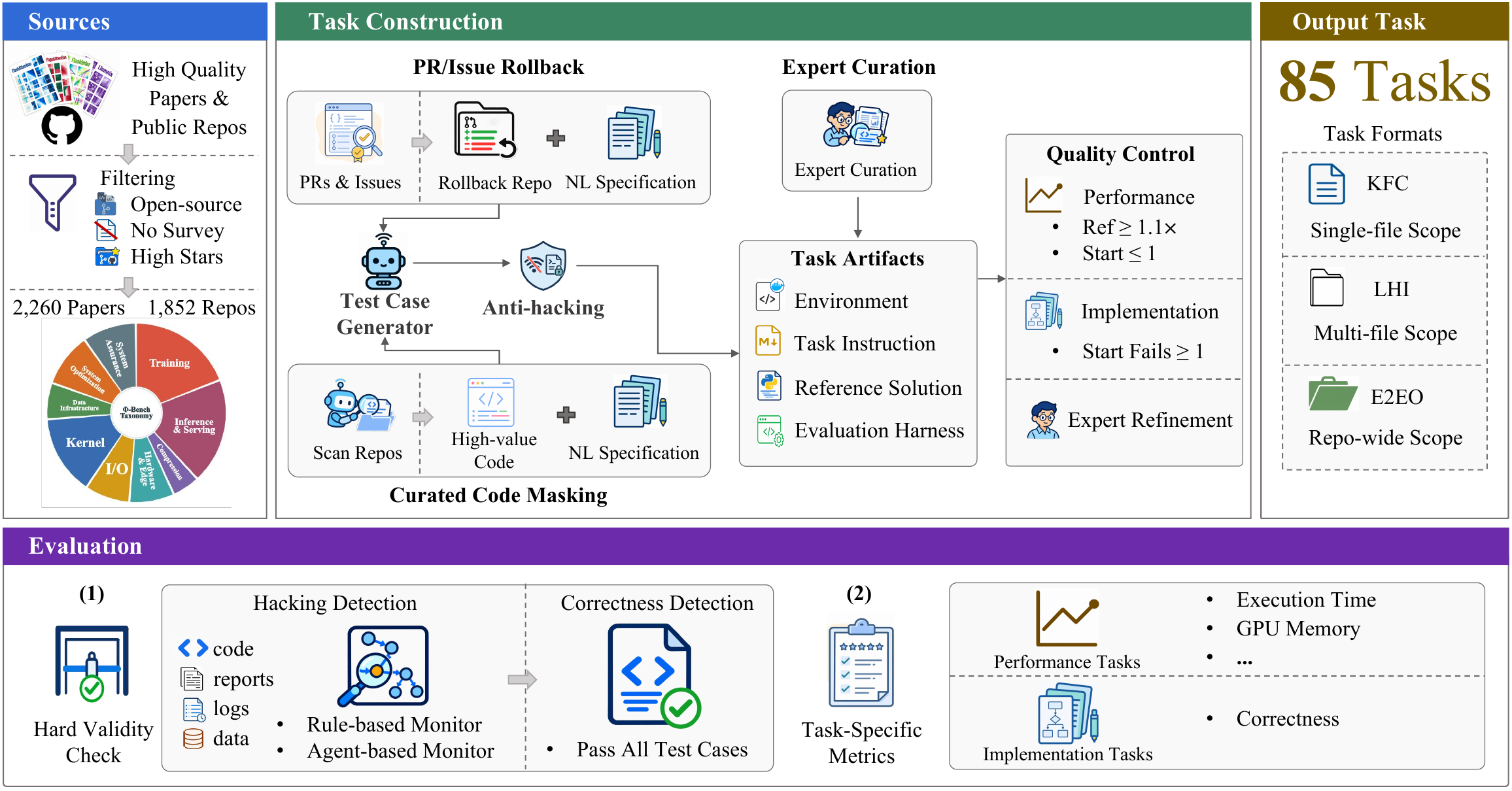}
    \caption{The task construction pipeline of $\Phi$-Bench.}
    \label{fig:pipeline}
    \vspace{-0.5em}
\end{figure*}

\section{Benchmark Design and Construction}
\label{sec:phi-bench}

In this section, we describe the task formats, task construction pipeline, and anti-hack based evaluation illustrated in Figure~\ref{fig:pipeline}.

\subsection{Task Formats}
\label{sec:task-format}


Each $\Phi$-Bench task comprises a natural-language (NL) specification, a LLM infrastructure repository codes, executable workloads or test cases, and an evaluation harness. All materials, including the test cases, are visible to the evaluated models except the evaluation harness. The specification defines either the required functionality or the performance objective. 

$\Phi$-Bench comprises three task formats with progressively broader edit scopes and less prescribed solution paths. KFC specifies a target function and restricts edits to one kernel file;  LHI tasks require agents to implement a repository-level feature over multiple steps and files; and E2EO only specifies an optimization objective and constraints while permitting repository-wide changes. Within each task format, tasks fall into one of two objectives: functional implementation or performance optimization. Representative examples of the three task formats are provided in Appendix~\ref{app:task-examples}.

\paragraph{Kernel Function Completion (KFC).}
A KFC task isolates a performance-critical kernel or operator whose interface and input-output semantics are explicitly specified. The agent completes or optimizes its implementation within a single file without changing the external interface. Submissions are first tested for functional correctness and numerical accuracy, after which correct solutions are benchmarked for efficiency. KFC therefore evaluates the implementation of correct and efficient computational primitives.

\paragraph{Long-Horizon Implementation (LHI).}
An LHI task provides an issue-style feature request and a coarse-grained editable scope, such as a repository submodule. Solving it requires navigating the codebase, understanding interactions among modules, modifying multiple files, and iteratively building, testing, and debugging the implementation. LHI thus evaluates nontrivial repository-level development rather than isolated function completion.

\paragraph{End-to-End Optimization (E2EO).}
An E2EO task provides a representative workload, a system-level optimization objective, and a set of constraints, without prescribing which components or strategies to use. The agent must profile the system, identify bottlenecks, formulate an optimization plan, and implement potentially repository-wide changes. E2EO therefore evaluates bottleneck discovery, cross-layer reasoning, and hypothesis-driven infrastructure optimization.

\subsection{Task construction Pipeline}
\label{sec:task-construction}

$\Phi$-Bench tasks are derived from published research and real-world engineering artifacts in LLM infrastructure. We draw on papers from systems and ML systems venues, including OSDI, SOSP, and MLSys, as well as issues, pull requests, and code from established open-source projects such as Megatron-LM~\citep{shoeybi2019megatron}, vLLM~\citep{kwon2023efficient}, and Verl~\citep{sheng2025hybridflow}. We analyze these sources to derive a coverage taxonomy that guides task construction.


\paragraph{Sourcing and filtering.}
Our sourcing combines recent systems research and widely used open-source LLM infrastructure repositories. For the research source pool, we collect papers published at leading systems conferences from 2023 to 2026. An agent-based screening pipeline retains papers that address the implementation, acceleration, or resource scheduling of LLM training or inference and provide publicly available code. For the engineering source pool, we inspect issues, pull requests(PR), and well-commented source code in production-oriented repositories. A second agent-based filter prioritizes artifacts that introduce substantial optimizations or new infrastructure capabilities, address hardware-specific limitations, or document challenging engineering problems. The resulting source pool comprises 2,260 papers and 1,852 repositories.

\paragraph{Coverage taxonomy construction.}

\begin{figure*}[!t]
    \centering
    \includegraphics[width=\textwidth]{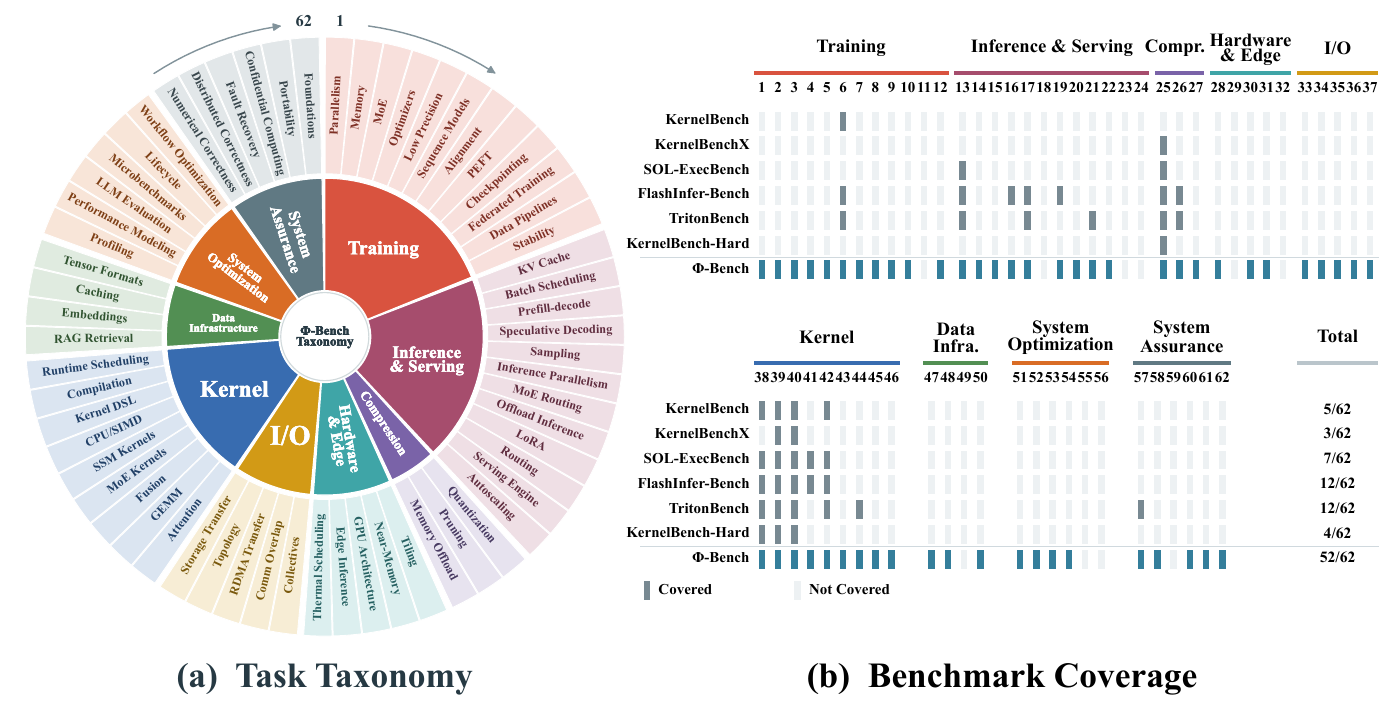}
    \caption{Task taxonomy and benchmark coverage. (a) 9 top-level categories comprising 62 middle-level topics. (b) Coverage of the same 62 topics by $\Phi$-Bench and related benchmarks. Filled cells indicate covered topics.}
    \label{fig:taxonomy}
\end{figure*}

We organize the selected sources into a hierarchical, three-level coverage taxonomy. For each source, we prompt a agent to assign labels at the three predefined levels. The top level captures broad areas of LLM infrastructure engineering and optimization. The middle level identifies more specific research problems, techniques, infrastructure components, and workload characteristics, while the bottom level consists of fine-grained tags describing the concrete problems addressed by the sources.


From the 4,112 papers and repositories in our source pool, we extract 1,163 task-relevant items, including curated codeS, pull requests, and optimization challenges. A labeling agent assigns fine-grained tags to these items, yielding 410 distinct tags as bottom level. We organize the 410 tags into 62 middle-level topics and nine top-level domains, and use the middle-level topics to guide task selection. Figure~\ref{fig:taxonomy} shows the taxonomy and compares $\Phi$-Bench's topic coverage with that of related benchmarks.

Guided by the coverage taxonomy, we construct tasks through three approaches: PR/Issue rollback, Curated code masking, and Expert curation. We describe each construction approach in detail below.

\paragraph{PR/Issue rollback.}
We identify important pull requests and issues from public LLM infrastructure repositories and reconstruct them as benchmark tasks. Each task preserves the original engineering objective, using the pre-change repository state as input and the post-change implementation as its reference solution. Localized, single-file changes typically yield KFC tasks, whereas changes spanning multiple files or components yield LHI or E2EO tasks.


\paragraph{Curated code masking.}
We employ agents to scan code repositories using a predefined construction pipeline and identify high-value implementation sites that can be converted into benchmark tasks. Depending on the scope and complexity of a selected site, we remove either a localized code segment or an entire implementation module.


\paragraph{Expert curation.}
For problems that cannot be reliably identified or reconstructed from repository changes alone, human experts curate challenging and high-value tasks from influential research papers and conference challenges. After identifying a target problem, the annotators construct an appropriate initial repository state by removing the relevant implementation, reverting the repository to a simpler baseline, or, when no mature implementation is available, formulating the task directly from the problem specification. They then manually design test cases and evaluation workloads that precisely exercise the targeted capability. This approach is primarily used for problems that require substantial infrastructure-level reasoning, involve open-ended solution strategies, or represent emerging challenges for which repository history does not provide a complete reference trajectory.

\subsection{Test Construction and Task Validation}

Human experts write task-specific correctness tests and design representative workloads for performance evaluation. The tests cover normal inputs, boundary conditions, error paths, and regression cases. For performance tasks, experts calibrate the measurement protocol to obtain stable measurements across repeated runs of the same implementation. To prevent agents from looking up solutions, we configure the sandbox's network proxy to an unreachable endpoint and provide a frozen repository snapshot without accessible Git history.

With these tests and workloads in place, a validation agent applies predefined checks to screen candidate tasks for validity and discriminative power. For performance-scored tasks, the reference solution must achieve a stable performance ratio of at least $1.1$ relative to the baseline in the official execution environment, while the starter or no-op solution must not outperform the baseline. For implementation-scored tasks, the reference solution must pass all tests in the same environment, while the starter solution must fail at least one. Candidates that fail these checks are revised or excluded. Human experts then review the validated candidates and select the final tasks for $\Phi$-Bench.

\subsection{Task Distribution}
\label{sec:task-distribution}

\begin{figure}[t]
    \centering
    \includegraphics[width=\linewidth]{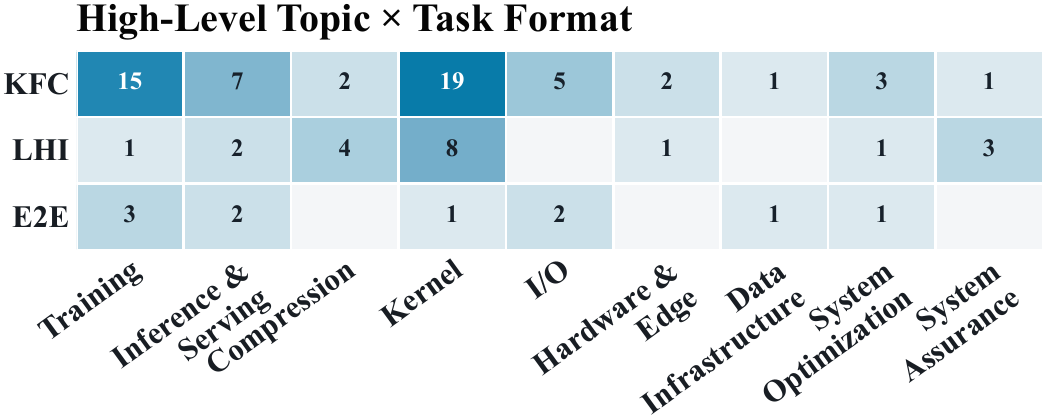}
    \caption{Task distribution in $\Phi$-Bench.}
    \label{fig:task}
\end{figure}

The final version of $\Phi$-Bench contains 85 tasks, comprising 55 KFC tasks, 20 LHI tasks, and 10 E2EO tasks. Figure~\ref{fig:task} summarizes their distribution across task formats and infrastructure topics. $\Phi$-Bench covers all nine top-level infrastructure domains, with multiple task formats represented across them.


\subsection{Evaluation Metrics}
\label{sec:metrics}

$\Phi$-Bench scores each task attempt according to its objective, independently of task format. Functional implementation tasks receive a binary reward in $\{0,1\}$, while performance optimization tasks receive a continuous reward in $[0,1]$; higher rewards indicate better outcomes. Both metrics are gated by checks for buildability, functional correctness, edit constraints, and reward hacking, all of which must be passed before a positive reward is assigned.
\begingroup
\setlength{\abovedisplayskip}{4pt}
\setlength{\belowdisplayskip}{4pt}
\setlength{\abovedisplayshortskip}{2pt}
\setlength{\belowdisplayshortskip}{3pt}
\paragraph{Performance Metric.}

For tasks with a performance objective, a submission must first pass hard validity checks for reward hacking, edit-scope compliance, and functional correctness. Let $t_i^{\mathrm{base}}$ and $t_i^{\mathrm{cand}}$ denote the baseline and candidate scores in paired measurement $i$, respectively. We define the candidate's relative performance as $s=\operatorname{median}_{i}(t_i^{\mathrm{cand}}/t_i^{\mathrm{base}})$.

We obtain the reference performance $s_{\mathrm{ref}}$ by repeatedly evaluating the reference solution in the same execution environment and taking the median performance. A candidate receives a reward of zero if the reference solution does not yield a valid performance ratio, i.e., $s_{\mathrm{ref}} \leq 1$. Otherwise, the reward is computed as

\[
R_{\mathrm{perf}} =
\begin{cases}
0, & s \leq s_{\mathrm{ref}},\\
\displaystyle
\min\!\left(
1,\;
\frac{\ln(s/s_{\mathrm{ref}})}{\ln s_{\mathrm{ref}}}
\right), & s > s_{\mathrm{ref}}.
\end{cases}
\]

This logarithmic normalization assigns a reward of zero when the candidate matches or underperforms the reference solution. For candidates that outperform the reference, the reward increases logarithmically and reaches its maximum value of $1.0$ when $s \geq s_{\mathrm{ref}}^{2}$.

\paragraph{Implementation Metric.}
For functional implementation tasks , we use the binary metric:

\[
R_{\mathrm{impl}} =
\begin{cases}
1, & \text{if all conditions and test cases pass},\\
0, & \text{otherwise}.
\end{cases}
\]

A candidate receives full credit only if it builds, imports, and passes all tests.
\endgroup
\paragraph{Reward-Hacking Detection.}
To preserve the integrity of $\Phi$-Bench, we use complementary rule-based monitoring and agent-based inspection. The rule-based monitor scans complete trajectories and final code changes for predefined violations, including searching GitHub for the original implementation, retrieving associated patches or commit diffs, and recovering upstream code from published Python packages. A dedicated proctor agent detects more complex hacking behaviors, including hard-coding expected outputs, bypassing correctness checks, and adding branches that artificially inflate measured performance. Any attempt flagged by either mechanism receives zero reward.



%% file: sections/4_evaluation.tex
\begin{table*}[t]
    \centering
    \begingroup
    \setlength{\tabcolsep}{2pt}
    \renewcommand{\arraystretch}{1.12}
    \caption{Category-wise performance on $\Phi$-Bench. Scores are
    reported as percentages, with the best result in each category
    highlighted in bold. All Models use the max reasoning effort.}
    \label{tab:performance}
    \footnotesize
    \begin{tabular*}{\textwidth}{
        @{\extracolsep{\fill}}
        l|cccccccccc
        @{}
    }
        \toprule
        Model
        & Training
        & \shortstack{Inference\\\& Serving}
        & Compression
        & Kernel
        & I/O
        & \shortstack{Hardware\\\& Edge}
        & \shortstack{Data\\Infrastructure}
        & \shortstack{System\\Optimization}
        & \shortstack{System\\Assurance}
        & Full \\
        \midrule

        Claude Opus 5
        & \textbf{56.40} & 26.20 & \textbf{15.70}
        & \textbf{30.00} & \textbf{57.10} & 3.90
        & \textbf{49.40} & 34.20 & 32.20 & \textbf{36.53} \\

        Kimi K3
        & 35.50 & \textbf{30.50} & 13.20 & 22.20 & 35.70
        & 3.60 & 27.40 & \textbf{41.10} & 39.10 & 28.12 \\

        Qwen3.8-Max
        & 43.90 & 29.90 & 8.30 & 23.00 & 18.30
        & 0.10 & 26.50 & 26.80 & 46.30 & 27.73 \\

        GPT 5.6 Sol
        & 39.90 & 21.80 & 3.00 & 17.70 & 38.50
        & 0.00 & 18.40 & 22.80 & 37.80 & 24.51 \\

        GLM 5.2
        & 28.70 & 19.20 & 9.00 & 18.60 & 19.10
        & 1.40 & 27.40 & 22.10 & \textbf{57.40} & 21.92 \\

        Claude Sonnet 5
        & 34.00 & 7.20 & 6.60 & 14.10 & 20.60
        & 0.00 & 23.50 & 1.50 & 33.70 & 17.58 \\

        Qwen3.7-Max
        & 27.40 & 12.70 & 5.90 & 13.20 & 17.40
        & \textbf{5.40} & 0.00 & 6.80 & 32.20 & 16.07 \\

        DeepSeek-V4-Pro
        & 27.80 & 3.90 & 0.50 & 13.10 & 3.70
        & 0.90 & 0.00 & 7.00 & 31.70 & 13.31 \\

        \bottomrule
    \end{tabular*}

    \endgroup
\end{table*}

\section{Evaluation}
\subsection{Experimental Setup}




\paragraph{Models \& Scaffolds.}
We evaluate eight models on $\Phi$-Bench: claude-opus-5~\citep{anthropic2026claudeopus5}, claude-sonnet-5~\citep{anthropic2026claudesonnet5}, gpt-5.6-sol~\citep{openai2026gpt56}, qwen3.8-max~\citep{qwen38}, qwen3.7-max~\citep{qwen37}, kimi-k3~\citep{moonshotai2026kimik3}, glm-5.2~\citep{glm5team2026glm5vibecodingagentic} and deepseek-v4-pro~\citep{xu2026deepseek}. For each model, we use its highest available reasoning effort and maximum supported context length. We evaluate gpt-5.6-sol with its default scaffold, Codex~\citep{openai2025codex}, and all other models with Claude Code~\citep{anthropic2026claudecode}.

\paragraph{Implementation Details.}

Each task is evaluated on one NVIDIA H20 GPU. KFC permits one formal submission per agent because repeated feedback on short and simple kernel workloads could allow agents to exploit cache reuse or persistent state, producing apparent speedups without improving the kernel. Performance optimization tasks in LHI and E2EO permit up to 16 submissions to support iterative refinement; the highest-scoring valid submission determines the task reward.

\subsection{Main Results}
Table~\ref{tab:performance} presents the performance of leading frontier models across the nine major categories of our benchmark. Even the strongest model, Claude Opus 5, achieves an overall score of only 36.53\%, followed by Kimi K3 at 28.12\% and Qwen3.8 Max at 27.73\%. These results indicate substantial room for improvement in real-world LLM infrastructure engineering.

Performance also varies considerably across infrastructure domains. Claude Opus 5 leads in five of the nine categories, demonstrating relatively broad competence, whereas other models exhibit more specialized strengths: Kimi K3 performs best on Inference \& Serving and System Optimization, Qwen3.7-Max on Hardware \& Edge, and GLM 5.2 on System Assurance. No model achieves uniformly strong performance across all categories. 
The Hardware \& Edge category appears to be particularly challenging for the models: even the best-performing model achieved only 5.4\%. This suggests that current frontier models still lack sufficient understanding of the hardware aspects of AI infrastructure and do not yet demonstrate strong engineering capabilities.

\subsection{Ablation Studies}

\begin{figure*}[t]
    \centering
    \includegraphics[width=\textwidth]{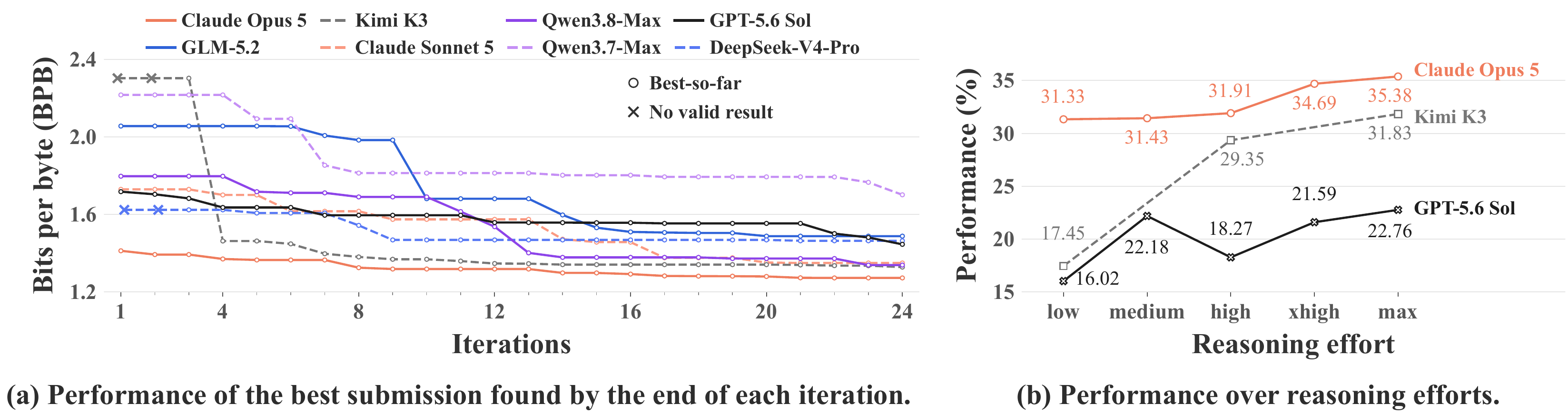}
    \caption{Ablation study on (a) iteration budget and (b) reasoning effort.}
    \label{fig:ablation}
    \label{fig:iteration}
    \label{fig:budget}
\end{figure*}


In this section, we analyze the impact of computational and interaction budgets on LLM systems engineering performance. Specifically, we investigate whether models improve through iterative refinement and whether increased test-time reasoning improves task performance.

\paragraph{Performance over iterations.}

To evaluate the models’ ability to iteratively refine their solutions, we run each model for 24 iterations on a challenging E2EO task from LLM training infrastructure. The task requires optimizing validation bits per byte (BPB) under a fixed compute budget and parameter constraint by jointly tuning token dispatch, load balancing, and routing configurations. Figure~\ref{fig:ablation} shows that most models exhibit some ability to improve through iteration. Claude Opus 5 achieves a strong initial BPB score and further improves over subsequent iterations, demonstrating effective long-horizon refinement. Qwen3.8-Max and Kimi K3, despite weak initial attempts, rapidly improve through iteration. In contrast, DeepSeek-V4-Pro, Qwen3.7-Max, GLM5.2, and GPT5.6 Sol exhibit prolonged plateaus and fail to reach comparable optimization levels.



\paragraph{Performance over reasoning budget.}

We evaluate reasoning effort for Claude Opus 5, Kimi K3, and GPT-5.6 Sol on the 20 LHI and 10 E2EO tasks, which involve broader repository context and multi-step work than KFC. All three models achieve their highest mean reward at maximum effort, with Kimi K3 showing the largest gain from low effort (14.38 \%; Figure~\ref{fig:ablation}). Higher effort does not always improve performance: GPT-5.6 Sol declines at the high setting relative to medium, whereas Opus 5 improves steadily.



%% file: sections/5_discussion.tex
\section{Analysis}
Beyond the main evaluation, we collect detailed solution trajectories from different models and analyze them to identify their predominant error modes and detect potential cheating or hacking behaviors. This trajectory-level analysis provides more fine-grained insights into current model limitations and informs future improvements to their capabilities.

\begin{figure*}[!t]
    \centering
    \begin{minipage}[t]{0.48\textwidth}
        \vspace{0pt}
        \centering
        \includegraphics[width=\linewidth]{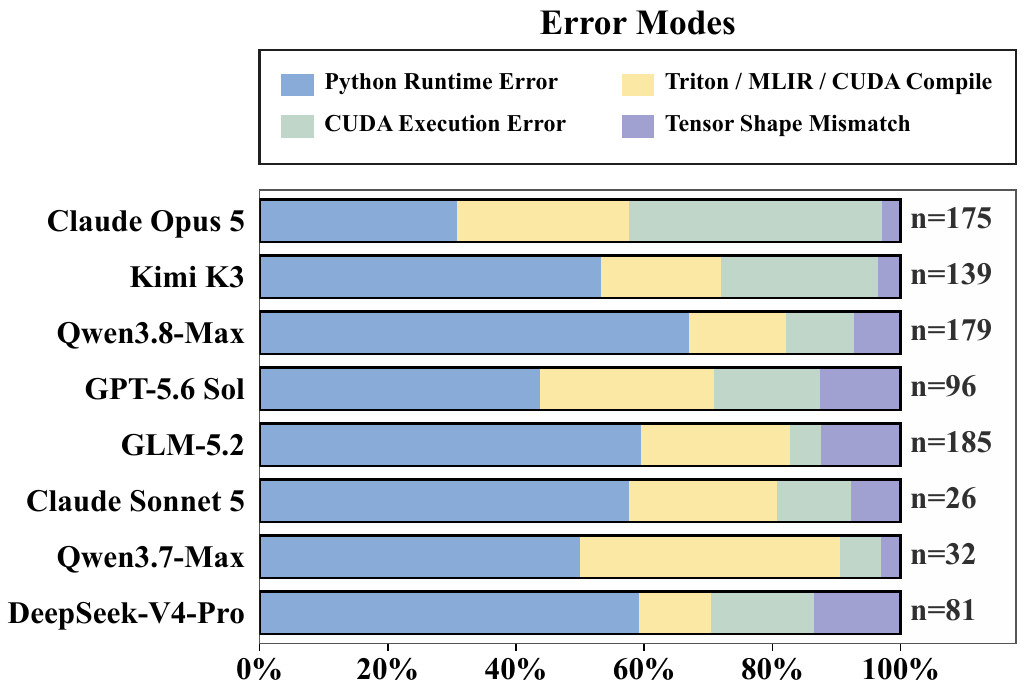}
        \captionof{figure}{Error mode breakdown.}
        \label{fig:error}
    \end{minipage}
    \hfill
    \begin{minipage}[t]{0.48\textwidth}
        \vspace{0pt}
        \centering
        \footnotesize
        \setlength{\tabcolsep}{8pt}
        \renewcommand{\arraystretch}{1.08}
        \captionof{table}{Performance across three task formats in $\Phi$-Bench. Scores are percentages, with the best result in each column highlighted in bold. Models are ordered by overall score.}
        \label{tab:performance_by_task_type}
        \begin{tabular}{l|cccc}
            \toprule
            Model & KFC & LHI & E2EO & Full \\
            \midrule
            Claude Opus 5  & \textbf{37.16} & \textbf{21.60} & \textbf{62.94} & \textbf{36.53} \\
            Kimi K3         & 26.09 & 19.55 & 56.41 & 28.12 \\
            Qwen3.8-Max     & 28.79 & 16.61 & 44.10 & 27.73 \\
            GPT 5.6 Sol     & 25.46 & 13.98 & 40.33 & 24.51 \\
            GLM 5.2         & 24.35 & 13.65 & 25.11 & 21.92 \\
            Claude Sonnet 5 & 18.14 & 13.46 & 22.74 & 17.58 \\
            Qwen3.7-Max     & 16.47 & 12.79 & 20.40 & 16.07 \\
            DeepSeek-V4-Pro  & 16.05 & 11.45 & 1.97 & 13.31 \\
            \bottomrule
        \end{tabular}
    \end{minipage}
    \vspace{-0.5em}
\end{figure*}

\subsection{Error Modes Analysis}

We collect error messages that appeared during each model's problem-solving trajectories and categorize them into four major classes: Python runtime error, CUDA execution error, Triton / MLIR / CUDA compile error, and tensor shape mismatch. Figure~\ref{fig:error} presents the number of errors and the distribution of error types across different models.

More capable models, including Claude Opus 5, Kimi K3, and Qwen3.8 Max, produce more errors, suggesting that they explore more solutions through iterative trial and correction. In contrast, lower-performing models generate fewer errors, indicating early termination or simpler solutions after limited attempts. This highlights that persistent exploration and iterative error correction are essential capabilities for solving challenging AI infrastructure engineering tasks. Error distributions further show that Claude Opus 5 has a lower proportion of Python runtime errors but more CUDA execution errors, suggesting that it quickly surpasses basic debugging stages and focuses more on low-level implementation and optimization.


\subsection{Capacity over task types. }

Table~\ref{tab:performance_by_task_type} compares model performance across three task categories with progressively increasing levels of open-endedness. Claude Opus 5 ranks first in every task format, achieving \(37.16\%\) on KFC, \(21.60\%\) on LHI, and \(62.94\%\) on E2EO, followed by Kimi K3 and Qwen3.8 Max. Compared with the less open-ended KFC and LHI categories, E2EO tasks show substantially larger score gaps across models. The strongest model, Claude Opus 5, scores 62.94, whereas the weakest, DeepSeek-V4-Pro, scores only 1.97, highlighting the strong discriminative power of highly open-ended tasks.

\subsection{Hacking Prevention \& Detection}


To mitigate hacking, we adopt two safeguards. First, we enforce soft network isolation by intercepting pip commands and requests to URLs associated with the evaluation repositories, blocking these requests and returning warning messages. Second, we define hacking behavior in the prompt and explicitly instruct the models to refrain from it. Across all evaluations, the rule-based detector flagged only three requests from DeepSeek-V4-Pro to retrieve code from the PyTorch website. The Proctor Agent reviewed these cases and found no actual hacking, indicating that these safeguards effectively prevented unauthorized code retrieval.

%% file: sections/7_case.tex
\section{Case Study}

To understand how models achieve effective long-horizon optimization of LLM infrastructure, we analyze the optimization trajectories of three representative models on the E2EO task in Figure~\ref{fig:iteration-current}: Claude Opus 5, the best-performing model on this task, and two weaker models, Qwen3.7-Max and DeepSeek-V4-Pro. Their iterative submissions reveal three capabilities that distinguish stronger models. First, they establish experimental foundations by building lightweight validation mechanisms to efficiently screen optimization directions. Second, they improve the quality of each iteration by controlling variables and measurement noise to extract reliable signals from experiments. Third, they cautiously attribute observed outcomes by considering alternative explanations and potential confounding factors. Detailed trajectory examples are provided in Appendix~\ref{sec:case-study}.

%% file: sections/6_conclusion.tex
\section{Discussion and Conclusion}

We introduced $\Phi$-Bench, a benchmark for evaluating whether LLMs can engineer the infrastructure that powers them. It contains 85 tasks grounded in real LLM training and inference codebases, covering all nine top-level categories and 52 of the 62 topics in our infrastructure taxonomy. The tasks come in three formats, ranging from kernel function completion to long-horizon implementation and repository-wide end-to-end optimization. They are produced by a construction pipeline that turns real PRs, issues, and repository code into executable tasks with validated tests, and every reward is gated by correctness and anti-hacking checks. Across eight frontier models, the strongest model scores only 36.53\%, and every model performs worse on long-horizon, repository-level implementation than on kernel-level tasks. The benefit of more reasoning effort or more refinement iterations also varies widely from model to model.

\begin{figure}[t]
    \centering
    \includegraphics[width=0.45\textwidth]{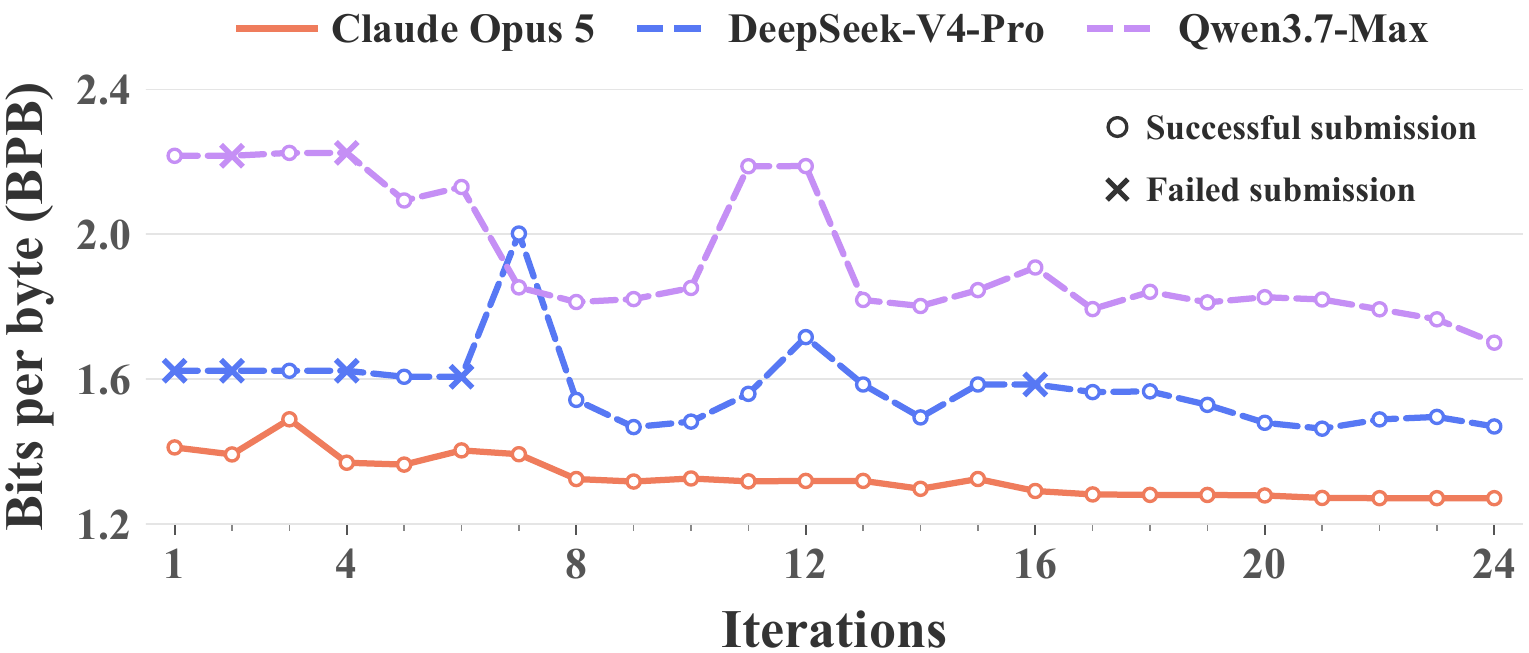}
    \caption{Performance of the current submission at each round.}
    \label{fig:iteration-current}
    
\end{figure}

\paragraph{Experimental discipline separates strong models.}
Our trajectory analysis (Appendix~\ref{sec:case-study}.) points to two habits beyond coding skill.
\emph{Validate cheaply before committing.} Claude Opus 5 builds small local proxies of the workload, first checks that they agree with the real measurement, and then uses them to discard weak ideas. Weaker models instead use formal submissions as their debugger, spending budget on failures that a quick local run would have caught.
\emph{Attribute cautiously.} When a change appeared to reduce throughput, Opus traced the drop to an invalidated compilation cache, reran the comparison under matched cache state, and kept the change, which in fact improved BPB. Weaker models more often credit or blame their latest edit directly.
Because infrastructure measurements are noisy and confounded by system state, these habits determine whether progress accumulates across iterations.

\paragraph{Implications and limitations.}
Building LLMs that can improve their own infrastructure will therefore require training and evaluating the experimental process itself, not only the correctness of the final code. This includes building proxies, controlling variables and noise, and ruling out confounders. Our trajectory analysis covers only one E2EO task, and quantifying these behaviors across more tasks and models is an important next step.

%% file: sections/8_appendix.tex
\section{Detailed Case Study}
\label{sec:case-study}

Using the task in Figure~\ref{fig:iteration}, we analyze the optimization trajectories of three representative models: Claude Opus 5, the best-performing model, and two weaker models, Qwen3.7-Max and DeepSeek-V4-Pro. Figure~\ref{fig:iteration-current} shows their per-round submissions. We ask a central question: \emph{what enables strong models to effectively perform iterative optimization of LLM infrastructure?} We identify three coupled capabilities: establishing experimental foundations, controlling variables and noise, and cautiously attributing outcomes.

\subsection{Experimental Foundations}

\textbf{Observations.}
Claude Opus 5 establishes lightweight local validation experiments to rapidly screen candidate hypotheses before formal submission. These experiments reproduce only the relevant part of the workload and serve as inexpensive local checks. For example, Opus compares a short local run against the development measurement and observes only a small BPB difference, confirming that the local experiment is sufficiently informative for preliminary screening. It then uses such checks to reject unpromising changes before committing to full evaluation.

These local checks form part of a coherent search strategy: Opus explores candidate directions broadly before focusing on the most promising ones.

Qwen3.7-Max and DeepSeek-V4-Pro, by contrast, largely follow an implementation--submission--observation loop without preliminary local validation. Formal submissions therefore serve simultaneously as debugging, hypothesis testing, and evaluation, increasing the cost of unsuccessful hypotheses. For example, DeepSeek applies \texttt{torch.compile} to expert modules without first validating checkpoint compatibility, eventually wasting a submission because the resulting checkpoint cannot be loaded.

\begin{takeawaybox}
\textbf{Takeaway.}
Good models establish low-cost mechanisms for hypothesis screening and validation, reducing the cost of each optimization iteration.
\end{takeawaybox}

\subsection{Variable and Noise Control}

\textbf{Observations.}
DeepSeek frequently modifies multiple variables simultaneously, making experimental outcomes difficult to attribute. Qwen3.7-Max exhibits the opposite failure mode: although it performs more disciplined single-variable tuning, many of its experiments produce differences comparable to measurement noise. Variable isolation alone is therefore insufficient when the observed effect is not reliably distinguishable.

Opus 5 explicitly considers both factors. It estimates measurement variation, accounts for differences in completed optimization steps, and focuses on experiments whose effects are sufficiently large to interpret. One representative transition changes the learning-rate schedule and worsens BPB from 1.3646 to 1.3932. Because the two runs use the same seed, complete the same number of steps, and have nearly identical runtime, Opus treats the degradation as reliable evidence rather than noise. This leads to the hypothesis that excessive time at very low learning rates is harmful, motivating the warmup--stable--decay schedule, which improves BPB to 1.3248. Rather than chasing each immediate improvement, Opus uses these results to update a consistent optimization plan.

\begin{takeawaybox}
\textbf{Takeaway.}
Good models conduct meaningful experiments and maximize the information gained from controlling variables and noise in each iteration.
\end{takeawaybox}

\subsection{Cautious Attribution}

\textbf{Observations.}
Opus 5 also differs in how cautiously it interprets experimental outcomes. Rather than directly attributing a performance change to its latest modification, it actively checks for alternative explanations and confounding factors.

For example, one candidate modification initially appears to cause a substantial throughput regression. Opus identifies that the code change invalidated the Inductor compilation cache, causing the run to incur additional cold-compilation overhead. It therefore repeats the comparison under matched cache conditions instead of rejecting the modification. The corrected experiment shows essentially unchanged throughput while retaining a BPB improvement. Similar cache-related artifacts are detected multiple times in its trajectory, with Opus repeatedly delaying attribution until an appropriate control is available. This protects its overall optimization plan from being redirected by a local measurement artifact. In contrast, Qwen3.7-Max and DeepSeek-V4-Pro more often interpret observed changes directly in terms of the latest intervention.

\begin{takeawaybox}
\textbf{Takeaway.}
Good models rigorously test alternative explanations before attributing experimental outcomes, preventing measurement artifacts from becoming incorrect optimization conclusions.
\end{takeawaybox}

Overall, these trajectories suggest that long-horizon infrastructure optimization requires more than reliable local execution. Claude Opus 5 maintains a coherent optimization strategy across rounds: it explores broadly, uses controlled comparisons to identify promising directions, and updates its plan based on reliable evidence. The weaker models more often make local, greedy decisions, making it difficult for useful knowledge to accumulate across iterations.

\section{Task Examples}
\label{app:task-examples}

We present one example from each of the three task formats: Mamba ZOH Discretization Operator, vLLM Multi-LoRA Serving, and vLLM Prefill Attention Optimization.

\subsection{KFC Example: Mamba ZOH Discretization Operator}
\label{app:kfc-example}

\paragraph{Task description.}
The task requires completing \texttt{discretize}, a missing function in
the preprocessing stage of Mamba's selective scan. Given continuous
state-space parameters and per-position step sizes, the function computes
the discrete state-transition factor and the input-dependent contribution.
The supplied \texttt{custom\_kernel} entry point invokes this function.

\paragraph{Environment and inputs.}
The execution environment provides a single NVIDIA H20 GPU and a Python
workspace with PyTorch and Triton. The inputs are FP32 CUDA tensors:
\texttt{u} and positive \texttt{delta} have shape \([B_t,L,D]\),
\texttt{A} has shape \([D,N]\), and \texttt{B} has shape \([B_t,L,N]\).
Here, \(B_t\), \(L\), \(D\), and \(N\) denote the batch size, sequence
length, channel count, and state dimension, respectively.
Both outputs have shape \([B_t,L,D,N]\), with entries
\(\texttt{deltaA}_{b\ell dn}=\exp(\delta_{b\ell d}A_{dn})\) and
\(\texttt{deltaB\_u}_{b\ell dn}=\delta_{b\ell d}B_{b\ell n}u_{b\ell d}\).
Each output entry can be computed independently.

\paragraph{Implementation constraints.}
The implementation must use FP32 arithmetic and return FP32 outputs
that match the reference with absolute and relative tolerances of
\(10^{-3}\). Correctness is required across single-timestep (\(L=1\)),
odd-sized, and non-power-of-two configurations. Outputs must be computed
from the supplied inputs. Retrieving or importing an external solution,
bypassing network isolation, and circumventing evaluation checks are
prohibited.

\paragraph{Submission.}
A single final implementation is submitted through the modified working
tree.
The task does not provide an iterative submission--feedback loop.
A submission receives a score of 1 only if all correctness and verifier
checks pass, and 0 otherwise.

\subsection{LHI Example: vLLM Multi-LoRA Serving}
\label{app:lhi-example}

\paragraph{Task specification.}
The task requires accelerating a functionally correct but inefficient
batched Multi-LoRA implementation in vLLM. The optimization target is
the complete shrink--expand computation: each token is projected through
the A matrix of its assigned adapter, scaled, and subsequently projected
through the corresponding B matrix into the designated output region.
Performance is evaluated for this batched adapter computation.
The agent must identify the relevant implementation and develop
optimizations that preserve its numerical behavior.

\paragraph{Environment and inputs.}
The provided code exposes the \texttt{lora\_shrink} and
\texttt{lora\_expand} operators in a single-GPU environment with vLLM,
PyTorch, and Triton. Inputs comprise token activations, LoRA A/B weights,
token-to-adapter mappings, and routing tensors that group tokens by
adapter. Workloads contain thousands of tokens assigned to multiple
low-rank adapters. Tokens with adapter identifier \(-1\) receive no
LoRA contribution.

\paragraph{Implementation constraints.}
The optimization scope is not restricted to a predefined set of files,
functions, or modules. Token activations and LoRA weights use FP16, with
FP32 accumulation, and outputs must satisfy the specified numerical
checks against the reference. Operator signatures, token-routing
semantics, in-place output writes, and the accumulation behavior of
\texttt{lora\_expand} must remain unchanged. Retrieving external reference
implementations, bypassing network isolation, and accessing the protected
evaluator or hidden workloads are prohibited.

\paragraph{Submission.}
The submission protocol permits one to sixteen rounds. Each round comprises an implementation
change, a submission, and a written analysis of the returned feedback
before further revision. Feedback includes correctness, development
speedup, the best result to date, and the remaining submission budget.
Early finalization requires a stated reason; otherwise, finalization
occurs automatically after the sixteenth submission. The candidate with
the highest development reward among those passing all development
checks is selected for final evaluation.

\subsection{E2EO Example: vLLM Prefill Attention Optimization}
\label{app:e2eo-example}

\paragraph{Task specification.}
The task requires optimizing variable-length causal prefill attention
on a single NVIDIA H20 GPU. Prompts of different lengths are packed
contiguously into a flat token tensor, with sequence boundaries specified
by \texttt{cu\_seqlens}. The objective is to reduce the latency of a
single attention call over the packed batch while preserving numerical
correctness.

\paragraph{Environment and inputs.}
The environment includes vLLM~0.10.1.1 with its Python implementation
accessible from the workspace, PyTorch, Triton, and a CUDA compiler.
A starting attention implementation and a public development benchmark
are provided. Inputs comprise contiguous BF16 query, key, and value
tensors, an int32 cumulative sequence-length array, and a preallocated
output buffer. The interface supports grouped-query attention, head
dimensions of 64 and 128, and arbitrary sequence lengths, including
empty and single-token sequences.

\paragraph{Implementation constraints.}
Repository-wide modifications are permitted, with no prescribed set of
files, functions, or modules to edit and no requirement to reuse vLLM's
existing attention kernels. The implementation must preserve the
\texttt{VarlenPrefillAttention} interface, including \texttt{prepare()}
and \texttt{forward()}; only \texttt{forward()} is timed.
Each query must attend to its complete causal prefix within the same
sequence, and outputs must match an FP32 reference within the specified
tolerance. The implementation must write every element of the supplied
output buffer and return that buffer without modifying the query, key,
or value tensors. Cached outputs and approximations that omit part of
the causal prefix are prohibited. Persistent workspace and transient
allocations during a timed call must jointly remain within 3\,GiB.
Access to protected evaluation files, interference with measurement,
and runtime Internet access are prohibited.

\paragraph{Submission.}
The agent submits candidate implementations iteratively, receiving
evaluation feedback after each round. In each subsequent round,
the implementation is refined based on previous results.
The process may terminate early, and the best-performing valid
candidate is selected for final evaluation.